\pdfoutput=1

\documentclass[11pt]{article}

\usepackage{acl}

\usepackage{latexsym}

\usepackage[T1]{fontenc}

\usepackage[utf8]{inputenc}

\usepackage{microtype}
\usepackage{times}  
\usepackage{helvet}  
\usepackage{courier}  
\usepackage{graphicx} 
\usepackage{natbib}  
\usepackage{caption} 
\DeclareCaptionStyle{ruled}{labelfont=normalfont,labelsep=colon,strut=off} 
\usepackage{algorithm}
\usepackage{algorithmic}
\usepackage{amssymb}
\usepackage{dsfont}
\usepackage{amsmath}
\usepackage{booktabs}
\usepackage{multirow}

\usepackage{newfloat}
\usepackage{listings}

\aclfinalcopy 

\title{HiCLRE: A Hierarchical Contrastive Learning Framework\\for Distantly Supervised Relation Extraction}

\author{Dongyang Li$^{1,2}$ \thanks{\ \ D. Li and T. Zhang contributed equally to this work.}, Taolin Zhang$^{3, 4}$\footnotemark[1], Nan Hu$^{1}$, Chengyu Wang$^{3}$, \textbf{Xiaofeng He}$^{1,5}$\thanks{\ \ Corresponding author.}\\
$^1$ School of Computer Science and Technology, East China Normal University \\
$^2$ Shanghai Key Laboratory of Trsustworthy Computing
$^3$ Alibaba Group \\
$^4$ School of Software Engineering, East China Normal University \\
$^5$ Shanghai Research Institute for Intelligent Autonomous Systems\\
 {\tt  dongyangli0612@gmail.com, zhangtl0519@gmail.com} \\
 {\tt hunan.vinny1997@gmail.com, chengyu.wcy@alibaba-inc.com} \\
 {\tt hexf@cs.ecnu.edu.cn}
 }

\begin{document}
\maketitle
\begin{abstract}

Distant supervision assumes that any sentence containing the same entity pairs reflects identical relationships.
Previous works of distantly supervised relation extraction (DSRE) task generally focus on sentence-level or bag-level de-noising techniques independently, neglecting the explicit interaction with cross levels.
In this paper, we propose a \textbf{Hi}erarchical \textbf{C}ontrastive \textbf{L}earning Framework for Distantly Supervised \textbf{R}elation \textbf{E}xtraction (HiCLRE) to reduce noisy sentences, which integrate the global structural information and local fine-grained interaction.
Specifically, we propose a three-level hierarchical learning framework to interact with cross levels, generating the de-noising context-aware representations via adapting the existing multi-head self-attention, named Multi-Granularity Recontextualization.
Meanwhile, pseudo positive samples are also provided in the specific level for contrastive learning via a dynamic gradient-based data augmentation strategy, named Dynamic Gradient Adversarial Perturbation.
Experiments demonstrate that HiCLRE significantly outperforms strong baselines in various mainstream DSRE datasets.\footnote{The source code and data can be available at \url{https://github.com/MatNLP/HiCLRE}}

\end{abstract}

\section{Introduction}
Relation extraction (RE) can draw relations of two entities from unstructured text. It can be widely used in natural language processing applications such as knowledge graph construction
\cite{DBLP:conf/aaai/KhatibHWJB020,DBLP:conf/acl/TangHWHZ20} and question answering \cite{DBLP:conf/acl/WangJ19,DBLP:conf/acl/LiuGFYCJLD20,DBLP:conf/acl/SaxenaTT20}.
Existing RE works \cite{DBLP:conf/acl/WeiSWTC20,DBLP:conf/acl/AltGH20,DBLP:conf/acl/VeysehDDN20} rely on a large-scale annotated dataset, which is time-consuming and labor-intensive.
DSRE \cite{DBLP:conf/acl/MintzBSJ09} attempts to address this issue via automatically generating training text samples.
Obviously, this assumption introduces noisy data and may hurt the performance.
Hence, multi-instance learning (MIL) \cite{DBLP:conf/emnlp/ZengLC015} is further proposed to assign a bag containing \textit{``at least one''} correct sentence of relation triple.
\begin{figure}
\centering
\includegraphics[height=5.5cm, width=7cm]{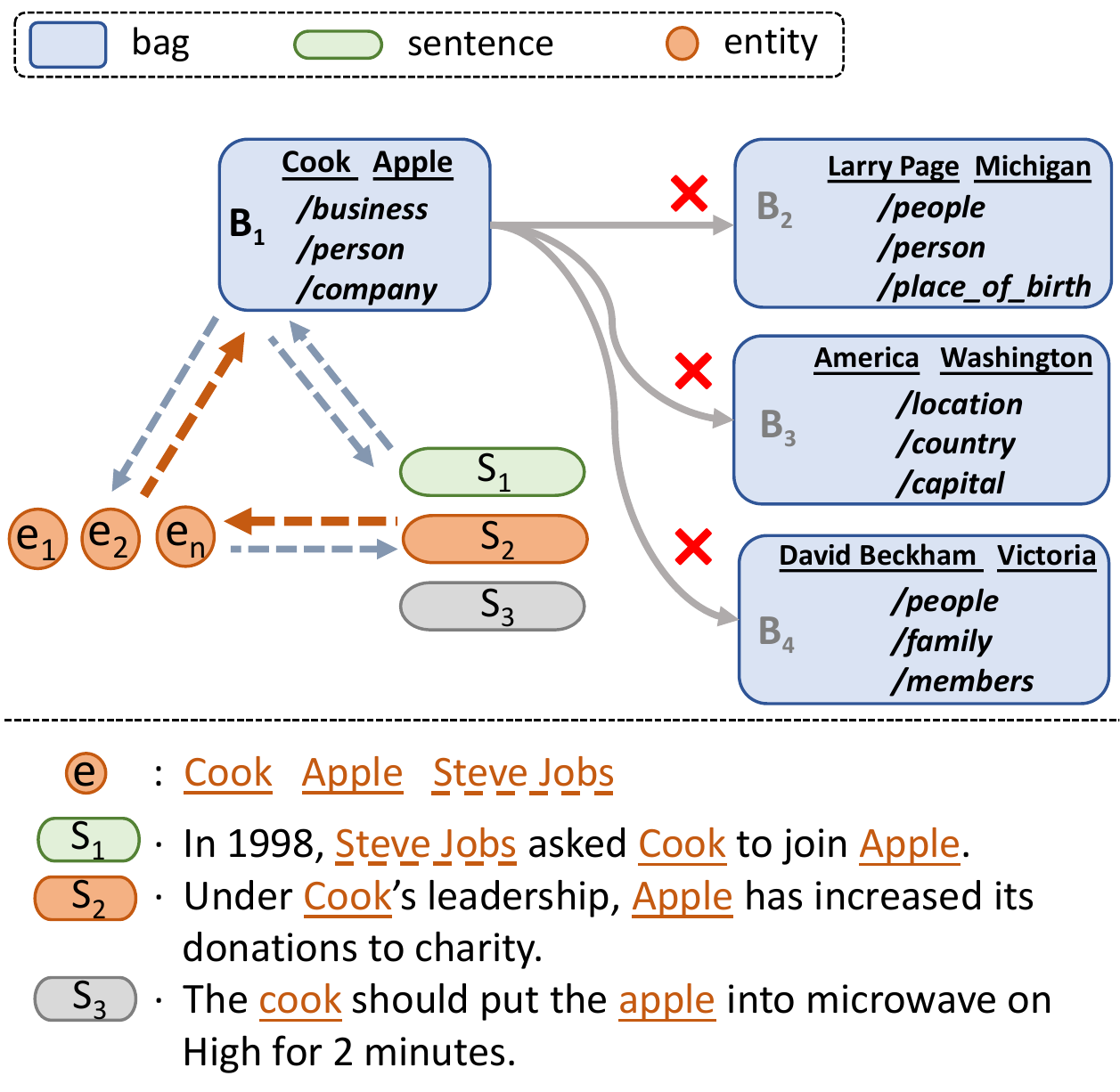} 
\caption{Example of semantic relationships in specific levels and cross levels. The red cross means the semantic difference of two bag-level relations and the dotted arrow indicates the semantic overlapping of cross levels. (Best viewed in color).}
\label{motivation_graph}
\end{figure}

The previous approaches of DSRE tackle the task at different granularities (i.e. sentence-level and bag-level).
(1) Sentence-level.
These works \citep{DBLP:conf/aaai/WuFZ19, DBLP:conf/naacl/LiZJZ19} focus on finding the ground-truth relational labels from the internal semantics of the input sentences.
(2) Bag-level.
Although these works \citep{DBLP:conf/ijcai/SuJ0ZL18,DBLP:conf/naacl/BeltagyLA19,DBLP:conf/acl/ChenST00Z20,DBLP:conf/naacl/ChristopoulouMA21} consider the information of sentence-level and bag-level simultaneously, but they ignore the explicit cross-level interactions, which contain plenty of knowledge to further boost the DSRE task performance.
As shown in Figure \ref{motivation_graph}, the rich semantic information of bag-level and sentence-level are provided for the ``Cook'' and ``Apple'' in the entity level.
For example, ``Steve Jobs'' in $s_1$ is also the co-founder of ``Apple'' company and the label of bag-level is the ``\textit{/business/person/company}'' exactly shows the relation of this entity pair.
Meanwhile, the huge semantic difference exists in a specific level such as the ``\textit{/business/person/company}'' and ``\textit{/location/country/capital}'' in bag level.

To overcome the challenges mentioned above, we propose a \textbf{Hi}erarchical \textbf{C}ontrastive \textbf{L}earning framework for distantly supervised \textbf{R}elation \textbf{E}xtraction (HiCLRE), which facilities semantic interactions within a specific level and cross levels:

\textbf{(1) Multi-Granularity Recontextualization:} To capture the cross-level structural information, we adapt the multi-head self-attention mechanism into three-level granularities, including entity-level, sentence-level and bag-level.
We align the context-aware feature of each layer with the input of attention mechanism respectively.
The refined representations as recontextualized interaction semantics are picked out for the corresponding level via the attention scores aggregated by the other two levels.

\textbf{(2) Dynamic Gradient Adversarial Perturbation:} To obtain the more accurate specific-level representations, we employ gradient-based contrastive learning \citep{DBLP:conf/cvpr/HadsellCL06,DBLP:journals/corr/abs-1807-03748} to pull the information of constructed pseudo positive samples and push the difference of negative samples.
Concretely, we calculate the dynamic perturbation from two aspects, including the normalized gradient of task loss and the temporal weighted memories similarity between the last and current epoch.
To verify the effectiveness of HiCLRE, we evaluate our model on three mainstream DSRE datasets, including NYT10 \citep{DBLP:conf/pkdd/RiedelYM10}, GDS \cite{DBLP:conf/akbc/JatKT17}, and KBP \cite{DBLP:conf/aaai/LingW12}.
The experimental results show that HiCLRE significantly outperforms the state-of-the-art baselines' performance, achieving a 2.2\% relative AUC increase and improving the P@M score from 77.2\% to 78.2\%.
Furthermore, the ablation study shows the individual contributions of each module.

Accordingly, the major contributions of this paper are summarized as follows:
\begin{itemize}
\item We propose a hierarchical contrastive learning framework for DSRE task (HiCLRE), which fully utilizes the semantic
interaction within the specific level and cross levels, reducing the influence of noisy data.
\item The multi-granularity recontextualization is proposed to enhance the cross-level interaction and the dynamic gradient adversarial perturbation learns better representations within three specific levels.
\item Extensive experiments show that our model outperforms the strong baseline over DSRE datasets and detailed analysis demonstrates the modules are also effective.
\end{itemize}

\section{Related Work}
\subsection{Distantly Supervised Relation Extraction}
Recently, these works are divided into two categories. (1) Human-designed Feature. \citep{DBLP:conf/emnlp/YaoHRM11} propose three types of LDA (i.e. Rel-LDA, Rel-LDA1, and Type-LDA) to cluster the similar triples together.
MIML \citep{DBLP:conf/acl/HoffmannZLZW11,DBLP:conf/emnlp/SurdeanuTNM12} and MIL \citep{DBLP:conf/emnlp/ZengLC015} attempt to relax the limitation of distantly supervision assumption to tackle the data generation problem.
(2) Neural Networks Representation. These models automatically generate the feature representation via end-to-end learning to reduce manual intervention.
\citep{DBLP:conf/acl/WangXQ18} introduce a generative adversarial training framework that provides a cleaned dataset for RE task.
\citep{DBLP:conf/naacl/YeL19} consider both inter-bag and intra-bag attention to handle the noise at sentence-level and bag-level independently. 
SENT \citep{DBLP:conf/acl/MaGLZHZ20} is a sentence-level framework to generate efficient training samples by negative training to filter the noisy data.
These works generally use the partial levels' information independently to explore the relational semantics.

\begin{figure*}[!htb]
\centering
\includegraphics[width=2.1\columnwidth]{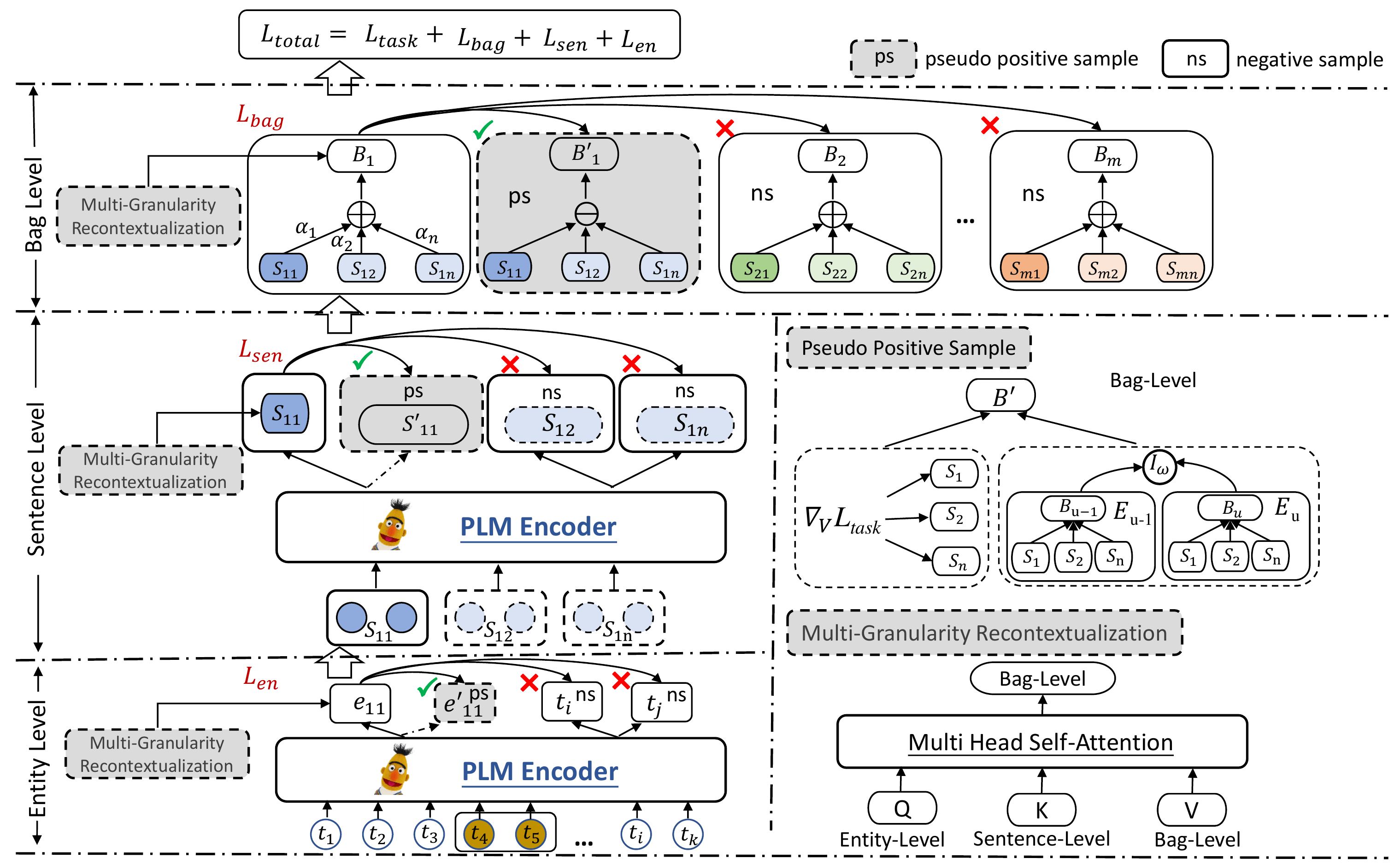} 
\caption{Model overview of HiCLRE. The left part is our model architecture and the right part shows the details on pseudo positive sample construction and multi-granularity recontextualization. (Best viewed in color).}.
\label{model_overview}
\end{figure*}

\subsection{Contrastive Learning}

\paragraph{Loss Function} 
NCE \citep{DBLP:journals/jmlr/GutmannH10} learns a classifier to distinguish the clean and noisy examples with the probability density function.
InfoNCE \citep{DBLP:journals/corr/abs-1807-03748} integrates the mutual information into the NCE, which can maximize similarity and minimize the difference.

\paragraph{Data Augmentation}

These works can be generally divided into three categories.
(1) Data augmentation by simple text processing. EDA \citep{DBLP:conf/emnlp/WeiZ19} proposes synonyms replace, randomly insert and randomly delete operations.
CIL \citep{DBLP:conf/acl/ChenST00Z20} utilizes TF-IDF scores to insert/substitute some unimportant words to/in instance to construct positive samples.
(2) Data augmentation by embedding processing. 
ConSERT \citep{DBLP:conf/acl/YanLWZWX20} explore four different data augmentation strategies (i.e. adversarial attack, token shuffling, cutoff and dropout) to generate views in BERT \citep{DBLP:conf/naacl/DevlinCLT19} embedding layer.
SimCSE \citep{DBLP:conf/emnlp/GaoYC21} applies twice dropout in the forward process to refine the better sentence representation.
(3) Data augmentation by external knowledge. ERICA \citep{DBLP:conf/acl/QinLT00JHS020} enumerates all the entity pairs in the training samples to link the corresponding relation from the external knowledge graph to obtain sufficient augmented data.
The mentioned above methods are generally augmenting from the data aspect, ignoring the influence of the changes inside the model during the training process \cite{DBLP:conf/acl/ZangQYLZLS20, DBLP:conf/acl/ZouHXDC20}.
Hence, we propose a hierarchical contrastive learning model to capture the global structure information and fine-grained interaction within the levels.


\section{Methodology}
\subsection{Model Overview and Notations}
The main architecture of our model is shown in Figure \ref{model_overview}.
The HiCLRE mainly includes two components. (1) Multi-Granularity Recontextualization aims to integrate the importance of cross levels to determine what valuable representation should be extracted in the target level. (2) Dynamic Gradient Adversarial Perturbation is proposed for specific levels to enhance the internal semantics via constructing the pseudo positive samples.

In HiCLRE, each sentence of input samples is consisted of certain tokens $S_{ij}=  \left ( t_{i1}, t_{i2},\cdots , t_{ik} \right )$, where $S_{ij}$ denotes the $i$-th sentence of bag $B_{j}$.
$k$ is the total number of tokens in $S_{ij}$ and $j$ represents the bag's index.
$e_{i1}$ and $e_{i2}$ are head and tail entity of sentence $S_{ij}$ respectively.
Each bag contains $n$ sentences $B_{j}= \left ( S_{1j},S_{2j},\cdots ,S_{nj} \right )$.
Our model aims to predict the specific relation $r_{j}$ of bag $B_{j}$ from $\left | r \right |$ relations.
$d$ denotes the hidden state dimension of pre-trained language models (PLMs).
\subsection{Hierarchical Learning Modeling}
We first introduce our hierarchical learning process including sentence-level and bag-level respectively and then describe the Multi-Granularity Recontextualization and Dynamic Gradient Adversarial Perturbation specifically.

\subsubsection{Sentence Representation}
To be specific, the input of sentence encoder is the token sequence of sentence $S_{ij}$ and its corresponding head entity $e_{i1}$ and tail entity $e_{i2}$\footnote{Entity's representation is calculated by averaging all tokens hidden states of the entity.}.
The textual encoder sums the token embedding, segment embedding and position embedding for each token to achieve its input embedding, and then computes context-aware hidden representations $H$= $\{h_{t_{i1}},h_{t_{i2}},\cdots, h_{e_{i1}},\cdots, h_{e_{i2}}, \cdots,h_{t_{ik}} \}$:
\begin{equation}
     H  = \mathcal{F}\left(\{t_{i1},t_{i2},\cdots , t_{ik}\}\right)
\end{equation}
where $\mathcal{F}$ is the PLMs (e.g. BERT) as our encoder and $H\in \mathbb{R}^{k\times d}$.
The sentence's embedding is calculated by the hidden representations of head entity, tail entity and the $[CLS]$ tag, which is in the first position of the input sequence to denote the whole semantic of the sentence.
\begin{equation}
h_{S_{ij}}= \sigma([h_{e_{i1}} \parallel h_{e_{i2}} \parallel h_{[CLS]}] \cdot W_{S})+b_{S}
\end{equation}
where the $\parallel$ means the concatenation operation, $W_{S}\in \mathbb{R}^{3d\times d }$ is a weight matrix and $b_{S}$ is the bias. $\sigma$ denotes the non-linear function.

\subsubsection{Bag Representation}
In this section, we use a sentence-level attention-based mechanism \citep{DBLP:conf/acl/LinSLLS16} to yield the aggregated bag representation.
Let $h_{B_{j}} \in \mathbb{R}^{ d}$ denotes the bag representation, and which is computed from the sentence's attention weight $\alpha _{ij}$ and hidden representation $h_{S_{ij}}$.
\begin{equation}
h_{B_{j}}= \sum_{i=1}^{n}\alpha _{ij} h_{S_{ij}} 
\end{equation}
To avoid naively treating each sentence of bags equally, the selective attention mechanism assigns the importance to reduce the noise instance.
Each weight $\alpha _{ij}$ is generated by a query-based function:
\begin{equation}
\alpha_{ij}=\frac{\exp \left(f_{ij}\right)}{\sum_{n} \exp \left(f_{ij}\right)}
\end{equation}
where $f_{ij}$ measures how well the input sentence $S_{ij}$ and the predicted relation $r_{j}$ matches.
\begin{equation}
f_{ij}=h_{S_{ij}} \mathbf{A_{j}} \mathbf{r_{j}}
\end{equation}
where $\mathbf{A_{j}}\in \mathbb{R}^{  d\times d}$ is a weighted diagonal matrix, and $\mathbf{r_{j}}\in \mathbb{R}^{d}$ is the representation of relation $r_{j}$ which is mapped from the relation label.
The final relation type of bag $B_{j}$ is predicted:
\begin{gather}
p(r_{j} \mid h_{B_{j}}, \theta)=\frac{\exp \left(O_{r}\right)}{\sum_{p=1}^{\left | r \right |} \exp \left(O_{p}\right)} \\
O_r= \sigma(W_r \cdot h_{B_{j}} ) + b_r
\end{gather}
where $W_r\in \mathbb{R}^{  \left | r \right |\times d }$ is trainable transformation matrix and $b_r \in \mathbb{R}^{ \left | r \right |}$ is the bias.
$\theta$ denotes bag encoder's parameters. $O_{r}\in \mathbb{R}^{  \left | r \right |}$ represents the final output of our model, which is associated with all relation types.
Therefore, the relation classification objective function of DSRE task is denoted as:
\begin{equation}
\mathcal{L}_{task}=-\sum_{j=1}^{\left | r \right |} \log p\left(r_{j} \mid h_{B_{j}}, \theta\right)
\end{equation}

\subsection{Multi-Granularity Recontextualization}
The hierarchical learning process described above neglects the explicit interaction of cross levels to refine the better level's representation.
Hence, after updating the hidden representations generated by the PLMs, our HiCLRE model attempts to recontextualize the enhanced representations for each level.
This is accomplished using a modified Transformer layer \citep{DBLP:conf/nips/VaswaniSPUJGKP17} that substitutes the multi-headed self-attention with multi-headed attention between the target level and the other two levels' representations.

Specifically, the underlying calculation process of multi-head self-attention is defined as:
\begin{equation}
\operatorname{Att.}(Q, K, V)=\operatorname{softmax}\left(\frac{Q K^{T}}{\sqrt{d_{k}}}\right) V
\end{equation}
where <$Q$, $K$, $V$> means query, key, and value respectively. $d_{k}$ is the dimension of $K$.
For example, if we focus on the enhanced bag-level representation \footnote{The calculation process of other modules is identical for entity level and sentence level. Hence, we take the bag level as an example in the following paper.}, the $h_{B_{j}}$ is substituted for the value, whereas the sentence-level $h_{S_{ij}}$ and entity-level $h_e$ mean the key and query respectively\footnote{Swapping the meaning of $Q$ and $K$ is also permitted.}:
\begin{equation}
    h_{B_{j}}^{'} = \operatorname{MLP}(\operatorname{Att.}(h_e, h_{S_{ij}}, h_{B_{j}}))
\end{equation}
where $\operatorname{MLP}$ is the linear multi-layer linear function.
The similarity calculation (i.e. query and key) acts as the cross-level information interaction attending to the bag-level representation.
After the interaction with multi-headed attention, we run a position-wise $\operatorname{MLP}$ similar to the standard transformer layer.
Next, we concatenate enhanced target level representation with original hierarchical hidden state to obtain an informative level's representation:
\begin{equation}
h_{B_{att_{j}}}= \sigma([h_{B_{j}} \parallel h_{B_{j}}^{'}] \cdot W_{att})+b_{att}
\end{equation}
where $W_{att}\in \mathbb{R}^{2d\times d }$ is a weight matrix and $b_{att}$ is the bias.
Finally, we leverage the three-level enhanced representation $h_{e_{att_{j}}}, h_{S_{att_{j}}}$ and $h_{B_{att_{j}}}$ to replace the hierarchical hidden representation in the following calculation process.


\subsection{Dynamic Gradient Adversarial Perturbation}
In addition to considering the interaction of cross levels, the semantic differences of fine-grained relations within the levels can also help models further enhance the context-aware representations.
We construct a pseudo positive sample for contrastive learning \citep{DBLP:journals/corr/abs-2011-00362} to push the dissimilar relations away.
Since the changes of specific-level gradient \cite{DBLP:journals/tist/ZhangSAL20} and the better context-aware semantic can boost the robustness representations, we devise the gradient perturbation and inertia weight memory mechanisms respectively.
\subsubsection{Gradient Perturbation}
\label{Gradient_Perturbation}
The continuous gradient perturbations $pt_{adv}$ is calculated from the gradient $g$ of the task loss with the parameter $V$.
\begin{equation}
g_{j}= \bigtriangledown _{V}\mathcal{L}_{task}(h_{B_{j}};\theta )
\end{equation}
where $V$ is the representation of the bag's sentences.
We differentiate the entity to generate the gradient perturbation for sentence level and the token for the entity level.

\begin{equation}
pt_{adv_{j}}= \epsilon \cdot \frac{g_{j}}{\left \| g_{j} \right \|}
\end{equation}
where $\left \| g \right \|$ is the norm of the gradient from the loss function, $ \epsilon$ is a hyperparameter to control the disturbing degree.

\subsubsection{Inertia Weight Memory}
With the training epoch increasing, we use the time-sequential information of different granularities to further improve the robustness of internal semantics.
Specifically, we add the inertia weight information \cite{DBLP:conf/eps/ShiE98} on the perturbation term, which takes advantage of the difference of representations between the last and the current epoch.
The inertia weight information is denoted as follows:
\begin{equation}
I_{w}=\frac{T-u}{T} \operatorname{sim}\left ( rep_{(u)},rep_{(u-1)} \right )
\end{equation}
where $T$ is the total epoch number of the training process and $u$ is the current epoch index. $rep_{(u)}$ can denote the entity, sentence, or bag representation respectively of the $u$-th epoch.
$rep$ is a embedding matrix saving the semantic memory in the order of element index, updated from the second epoch during the training process.
Then, we combine the inertia weight information with gradient perturbation for bag level:
\begin{equation}
pt_{{adv}_{j}}=\epsilon \frac{g_j}{\left \| g_j \right \|} + \frac{T-u}{T} \operatorname{sim}\left ( rep_{(u)},rep_{(u-1)} \right )
\end{equation}

We add $pt_{adv_{j}}$ into the bag embedding, and get pseudo positive sample $ h^{'}_{B_{j}}=h_{B_{j}}+ pt_{adv_{j}}$.
Then we randomly sample a bag in the batch act as the negative sample.
The positive and negative samples in InfoNCE loss \citep{DBLP:journals/corr/abs-1807-03748} are replaced by the dynamic gradient perturbations and random bags respectively:
\begin{equation}
\small
\mathcal{L}_{bag}^{info}=-\log \frac{\exp \left(cos \left(h_{B_{j}}, h^{'}_{{B_{j}}}\right) / \tau\right)}{\sum_{k=1}^{m} \mathds{1}_{[k \neq j]} \exp \left(cos \left(h_{B_{j}}, h_{B_{kj}}\right) / \tau\right)}
\end{equation}
where $\mathds{1}_{[k \neq j]}$ is an indicator function, $\tau$ is a hyper-parameter and $cos$ is the cosine function.
Due to the different granularities in the hierarchical framework, we devise different memories for entity-level, sentence-level, and bag-level, respectively.


\subsection{Training Objective}
In HiCLRE, our training objective contains two components, including the DSRE task loss and the contrastive learning loss.
The total loss of contrastive learning is the sum of three-level infoNCE loss.
Therefore, the overall objective function is formulated as follows:
\begin{equation}
\mathcal{L}_{total} = \lambda _{1}\mathcal{L}^{info}_{en}+ \lambda _{2}\mathcal{L}^{info}_{sen}+ \lambda _{3}\mathcal{L}^{info}_{bag}
   + \lambda _{4}\mathcal{L}_{task}
\end{equation}
where $\lambda _{l}$ is hyper-parameter and $\quad \sum_{l=1}^{4}\lambda_{l}=1$, denoting the weight of each components.

\begin{table*}[ht]
\small
\centering
\begin{tabular}{c|ccccc|ccccc}
\toprule
                         & \multicolumn{5}{c|}{NYT10}                                                                                                                                         & \multicolumn{5}{c}{GDS}                                                                                                                                                                                                                                                              \\ \cmidrule(r){2-11} 
\multirow{-2}{*}{Models} & \multicolumn{1}{c|}{AUC}            & \multicolumn{1}{c|}{P@100}         & \multicolumn{1}{c|}{P@200}         & \multicolumn{1}{c|}{P@300}         & P@M           & \multicolumn{1}{c|}{AUC}                                  & \multicolumn{1}{c|}{P@500}                                & \multicolumn{1}{c|}{P@1000}                               & \multicolumn{1}{c|}{P@300}                                & P@M                                  \\ \midrule
Mintz                    & \multicolumn{1}{c|}{10.7}           & \multicolumn{1}{c|}{52.3}          & \multicolumn{1}{c|}{50.2}          & \multicolumn{1}{c|}{45.0}          & 49.2          & \multicolumn{1}{c|}{-}                                    & \multicolumn{1}{c|}{-}                                    & \multicolumn{1}{c|}{-}                                    & \multicolumn{1}{c|}{-}                                    & -                                    \\
PCNN-ATT                 & \multicolumn{1}{c|}{34.1}           & \multicolumn{1}{c|}{73.0}          & \multicolumn{1}{c|}{68.0}          & \multicolumn{1}{c|}{67.3}          & 69.4          & \multicolumn{1}{c|}{79.9}                                 & \multicolumn{1}{c|}{90.6}                                 & \multicolumn{1}{c|}{87.6}                                 & \multicolumn{1}{c|}{75.2}                                 & 84.5                                 \\
MTB-MIL                  & \multicolumn{1}{c|}{40.8}           & \multicolumn{1}{c|}{76.2}          & \multicolumn{1}{c|}{71.1}          & \multicolumn{1}{c|}{69.4}          & 72.2          & \multicolumn{1}{c|}{88.5}                                 & \multicolumn{1}{c|}{94.8}                                 & \multicolumn{1}{c|}{92.2}                                 & \multicolumn{1}{c|}{87.0}                                 & 91.3                                 \\
RESIDE                   & \multicolumn{1}{c|}{41.5}           & \multicolumn{1}{c|}{\underline{81.8}}          & \multicolumn{1}{c|}{75.4}          & \multicolumn{1}{c|}{\underline {74.3}}    & {\underline {77.2}}    & \multicolumn{1}{c|}{89.1}                                 & \multicolumn{1}{c|}{94.8}                                 & \multicolumn{1}{c|}{91.1}                                 & \multicolumn{1}{c|}{82.7}                                 & 89.5                                 \\
REDSandT                 & \multicolumn{1}{c|}{42.4}           & \multicolumn{1}{c|}{78.8}          & \multicolumn{1}{c|}{75.0}          & \multicolumn{1}{c|}{73.0}          & 75.3          & \multicolumn{1}{c|}{86.1}                                 & \multicolumn{1}{c|}{95.6}                                 & \multicolumn{1}{c|}{92.6}                                 & \multicolumn{1}{c|}{84.6}                                 & 91.0                                 \\
DISTRE                   & \multicolumn{1}{c|}{42.2}           & \multicolumn{1}{c|}{68.0}          & \multicolumn{1}{c|}{67.0}          & \multicolumn{1}{c|}{65.3}          & 66.8          & \multicolumn{1}{c|}{89.9}                                 & \multicolumn{1}{c|}{97.0}                                 & \multicolumn{1}{c|}{93.8}                                 & \multicolumn{1}{c|}{87.6}                                 & 92.8                                 \\
CIL                      & \multicolumn{1}{c|}{\underline {43.1}}     & \multicolumn{1}{c|}{ 81.5}    & \multicolumn{1}{c|}{\underline {75.5}}    & \multicolumn{1}{c|}{72.1}          & 76.9          & \multicolumn{1}{c|}{\underline {90.8}}                           & \multicolumn{1}{c|}{\underline {97.1}}                           & \multicolumn{1}{c|}{\underline {94.0}}                           & \multicolumn{1}{c|}{\underline {87.8}}                           & {\underline {93.0}}                           \\ \midrule
HiCLRE(ours)              & \multicolumn{1}{c|}{\textbf{45.3}} & \multicolumn{1}{c|}{\textbf{82.0}} & \multicolumn{1}{c|}{\textbf{78.5}} & \multicolumn{1}{c|}{74.0} & \textbf{78.2} & \multicolumn{1}{c|}{{\textbf{95.5}}} & \multicolumn{1}{c|}{{ \textbf{99.6}}} & \multicolumn{1}{c|}{{ \textbf{98.4}}} & \multicolumn{1}{c|}{{ \textbf{98.3}}} & {\textbf{98.8}} \\ \bottomrule
\end{tabular}
\caption{General experimental results of HiCLRE and baselines on NYT10 and GDS datasets.}
\label{general_results}
\end{table*}

\section{Experiments}

\subsection{Datasets and Baselines}
We evaluate our HiCLRE model on three DSRE datasets, including NYT10 \citep{DBLP:conf/pkdd/RiedelYM10}, GDS \cite{DBLP:conf/akbc/JatKT17}, and KBP \cite{DBLP:conf/aaai/LingW12}.
Table \ref{dataset_details} shows the detailed statistics.
NYT10 is annotated from the New York Times and aligned to Freebase and NYT10-M removes the noisy relation types manually from NYT10.
GDS is extracted from human-judged Google Relation Extraction corpus.
KBP is constructed over the newswire and web text from the corpus, which is used in the yearly TAC Knowledge Base Population challenges \cite{ji2010overview}. 
Statistics of four datasets are showed in Appendix \ref{appendix_Datasets_Statistics}.

Mintz \citep{DBLP:conf/acl/MintzBSJ09} concatenates various features of sentences to train a multi-class logistic regression classifier.
PCNN-ATT \citep{DBLP:conf/acl/LinSLLS16} proposes a selective attention-based piece-wise CNN to get sentence embeddings.
MTB-MIL \cite{DBLP:conf/acl/SoaresFLK19} proposes a Matching the Blanks method to learn the sentences' representation by the entity linked text.
RESIDE \cite{DBLP:conf/emnlp/VashishthJPBT18} exploits the information of entity type and relation alias to add a soft limitation for relation classification.
REDSandT \citep{DBLP:journals/access/ChristouT21} employs the PLMs to focus on instance embedding, aggregating the representations to the attention modules.
DISTRE \citep{DBLP:conf/acl/AltHH19} combines the selective attention to its Transformer-based model.
CIL \citep{DBLP:conf/acl/ChenST00Z20} proposes a contrastive instance learning method under the MIL framework.

\subsection{Evaluation Metrics}
Following the previous works \citep{DBLP:conf/aaai/ChenSLTSCZ21}, we adopt the five general evaluation metrics in DSRE task to evaluate the performance, including \textbf{AUC}, \textbf{P@N} and \textbf{P@M}.
Specifically, AUC (i.e. Area Under Curve) depicts the area under the ROC curve \footnote{ROC curve is plotted by false positive rate and true positive rate.}.
P@N refers to the P@100, P@200 and P@300 used in the metrics, denoting the top 100, top 200 and top 300 precision respectively.
P@M is the mean value of the above three P@N results.

\subsection{Parameter Settings}
The underlying encoders of the entity level and sentence level are implemented by BERT\_base \citep{DBLP:conf/naacl/DevlinCLT19}.
The backbone encoder contains 12 Transformer layers and 12 self-attention heads, generating 768 hidden units for each token context-aware representation.
During the training stage, we set the model's learning rate as \{1e-5, 2e-5, 2e-7\}.
We choose AdamW \citep{DBLP:journals/corr/abs-1711-05101} as our model's loss optimizer, which weight decay is 1e-5 and learning rate is 0.1.
The max epoch is set to 5.
We find the best hyper-parameter of temperature $\tau$ is 0.05, the $\lambda$ set is \{0.4, 0.4, 0.1, 0.1\} and $\epsilon$ is 2. We show the important hyper parameters' searching results at Appendix \ref{appendix_parameters_commissioning_process}.


\subsection{General Experimental Results}
We first evaluate our HiCLRE model in the NYT10 and GDS that are popular used datasets in the DSRE task.
Table \ref{general_results} shows the overall performance on the NYT10 and GDS datasets.
From the results\footnote{CIL is the SOTA model in these datasets, whereas the source code is not provided so far. Hence, we reproduce the CIL model and report the performance in the test set.}, we can observe that (1) On both two datasets, the performance of our HiCLRE model outperforms all the strong baseline models significantly on the four metrics, achieving a new state-of-the-art result. (2) The performance of HiCLRE is greatly improved compared with the strongest baseline in two distantly supervised datasets (i.e. +2.2 AUC / +4.7 AUC). Meanwhile, we find the results of other four metrics are also increasing consistently.
In general, it can be seen from Table \ref{general_results} that the multi-granularity recontextualization for cross levels interaction and the dynamic gradient-based adversarial perturbation for specific levels can improve the performance greatly. 
Some general cases also prove the effectiveness in Appendix \ref{appendix_Case_Study}.

\begin{figure}[h]
\centering
\includegraphics[width=0.9\columnwidth]{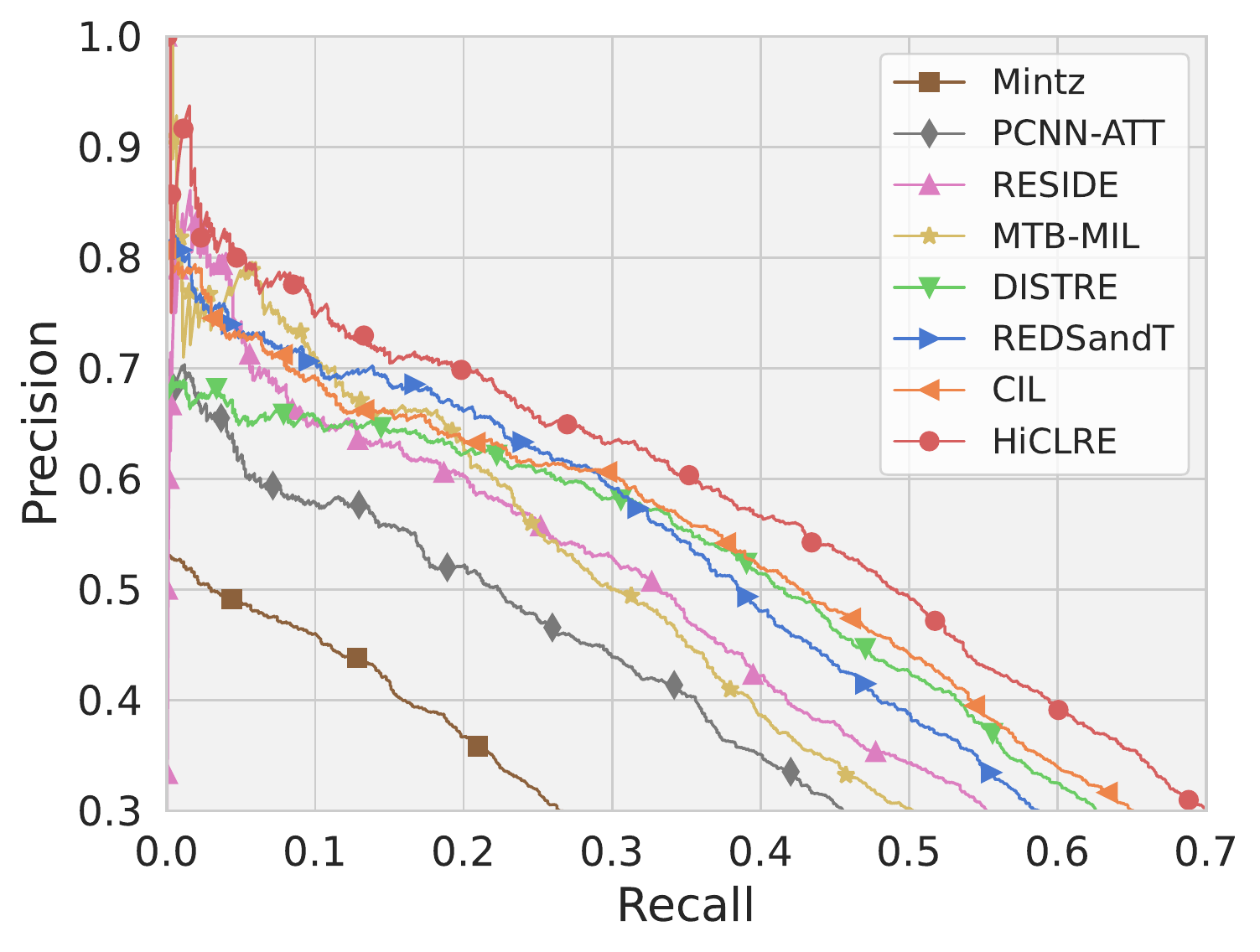} 
\caption{PR-curves of HiCLRE and other baselines on NYT10 dataset. (Best viewed in color)}
\label{pr_curves_nyt10}
\end{figure}

The baselines and HiCLRE's overall PR-cureve is illustrated in Figure \ref{pr_curves_nyt10}.
From the curve, we can observe that (1) Our HiCLRE shows higher precision and recall results compared to other strong baselines. (2) Although the curve initially fluctuates quite a bit, both metrics of HiCLRE are basically stabilized at a relatively large gap during the training process.
We conjecture that the difference of hierarchical context-aware representation is not obvious at the beginning of the model’s training and the stored representations for inertia weight memory of specific levels do not exist in the first training epoch.
During the training process, the mentioned above two learning problems tend to be stable and the performance of these two metrics is continuously performing better.

\begin{table}[!h]
\small
\begin{tabular}{c|ccc|ccc}
\toprule
\multicolumn{1}{c|}{\multirow{2}{*}{Models}} & \multicolumn{3}{c|}{NYT10-M}                                                                 & \multicolumn{3}{c}{KBP}                                                                           \\ \cmidrule(r){2-7} 
\multicolumn{1}{c|}{}                        & \multicolumn{1}{c|}{AUC}        & \multicolumn{1}{c|}{F1}         & \multicolumn{1}{c|}{P@M} & \multicolumn{1}{c|}{AUC}           & \multicolumn{1}{c|}{F1}            & \multicolumn{1}{c}{P@M} \\ \midrule
PCNN-A                                       & \multicolumn{1}{l|}{41.9}       & \multicolumn{1}{l|}{32.0}       & 68.6                     & \multicolumn{1}{l|}{15.4}          & \multicolumn{1}{l|}{31.5}          & 32.8                    \\
DISTRE                                       & \multicolumn{1}{l|}{35.7}       & \multicolumn{1}{l|}{31.4}       & 65.1                     & \multicolumn{1}{l|}{22.1}          & \multicolumn{1}{l|}{37.5}          & 46.4                    \\
CIL                                          & \multicolumn{1}{l|}{{\underline{56.0} }} & \multicolumn{1}{l|}{{\underline{34.3} }} & {\underline{75.9} }               & \multicolumn{1}{l|}{{\underline{29.5} }}    & \multicolumn{1}{l|}{{\underline{41.6} }}    & {\underline{47.3} }              \\ \midrule
HiCLRE                                 & \multicolumn{1}{l|}{\textbf{61.4}}           & \multicolumn{1}{l|}{\textbf{36.9}}           &         \multicolumn{1}{l|}{\textbf{88.0}}                  & \multicolumn{1}{l|}{\textbf{46.1}} & \multicolumn{1}{l|}{\textbf{61.0}} & \textbf{56.4}           \\ \bottomrule
\end{tabular}
\caption{Experiment results on human-annotated datasets.}
\vspace{-1em}
\label{human_annotated_dataset_experiments}
\end{table}

\subsection{Evaluation on Human Annotated Dataset}
Due to the inevitable annotated errors of distantly supervision assumption, we further evaluate our model on the human-annotated high-quality relation extraction datasets, including NYT10-M and KBP. The performances of baselines and HiCLRE are shown in Table \ref{human_annotated_dataset_experiments}.
The result shows that HiCLRE can significantly outperform the three strong baselines especially the AUC metric reaches 46.1, which improves about 50\% ($29.5 \mapsto 46.1$) performance than CIL \citep{DBLP:conf/acl/ChenST00Z20} on the KBP dataset.
This phenomenon implies that our model possesses a steady generalization ability to other analogous relation extraction datasets.

\section{Detailed Analysis of HiCLRE}
\subsection{Ablation Study}
To verify the effectiveness of various modules in our HiCLRE model, we conduct ablation study experiments on the NYT10 dataset. 
Specifically, we remove the following argued contributions in turn to evaluate the performance, including multi-granularity recontextualization, three-level contrastive learning loss, and the data augmentation strategies in each level.
The final results are shown in Table \ref{Ablation Study}.
From the results, we conclude that (1) The context-aware representation interactions for cross levels and the enhanced internal semantics representations for specific level are essential, dropping -1.8\% and -2.7\% point on the AUC metric respectively.
(2) We also find the sentence-level data augmentation skills for our HiCLRE model are the most important (e.g. -4.9\% and -2.6\% on AUC) compared to the other two levels.
The possible reason may be that the sentence granularity is the fundamental input granularity for the DSRE task including the term ``bag'' is also constructed by choosing the sentences with identical entity pairs.

\begin{table}[!ht]
\small 
\centering
\begin{tabular}{cc|c|c|c}
\toprule
\multicolumn{2}{c|}{Methods}                                          & \multicolumn{1}{c|}{AUC}   & \multicolumn{1}{c|}{F1}   & P@M                  \\ \midrule
\multicolumn{2}{c|}{HiCLRE}                                            & \multicolumn{1}{c|}{45.3} & \multicolumn{1}{c|}{49.5} & 78.2                 \\ \midrule
\multicolumn{2}{c|}{-Multi-Gra. Recon.}                                   & \multicolumn{1}{c|}{43.5}  & \multicolumn{1}{c|}{47.1} & 76.4                 \\
\multicolumn{2}{c|}{-Three-level CL Loss}                                   & 42.6                       & 47.9                      & 70.8                 \\ \midrule
\multicolumn{1}{l|}{\multirow{2}{*}{-Bag Level}}      & -bag gradient & 43.9                       & 48.1                      & 75.4                 \\
\multicolumn{1}{l|}{}                                 & -bag memory   & 42.4                       & 48.6                      & 72.3                 \\ \midrule
\multicolumn{1}{l|}{\multirow{2}{*}{-Sentence Level}} & -sen. gradient & 40.4                       & 48.9                      & 73.9                 \\
\multicolumn{1}{l|}{}                                 & -sen. memory   & \multicolumn{1}{c|}{42.7}  & \multicolumn{1}{c|}{48.1} & 67.2                 \\ \midrule
\multicolumn{1}{l|}{\multirow{2}{*}{-Entity Level}}   & -en. gradient  & 43.0                       & 48.4                      & 70.1                 \\
\multicolumn{1}{l|}{}                                 & -en. memory    & \multicolumn{1}{c|}{43.1}      & \multicolumn{1}{c|}{46.7}     & \multicolumn{1}{c}{73.2} \\ \bottomrule
\end{tabular}
\caption{Ablation study of HiCLRE on NYT10. ''-`` means  removing the module behind. }
\label{Ablation Study}
\end{table}

\subsection{The Influence of Multi-Granularity Recontextualization}
\begin{figure}[h]
\centering
\includegraphics[width=0.9\columnwidth]{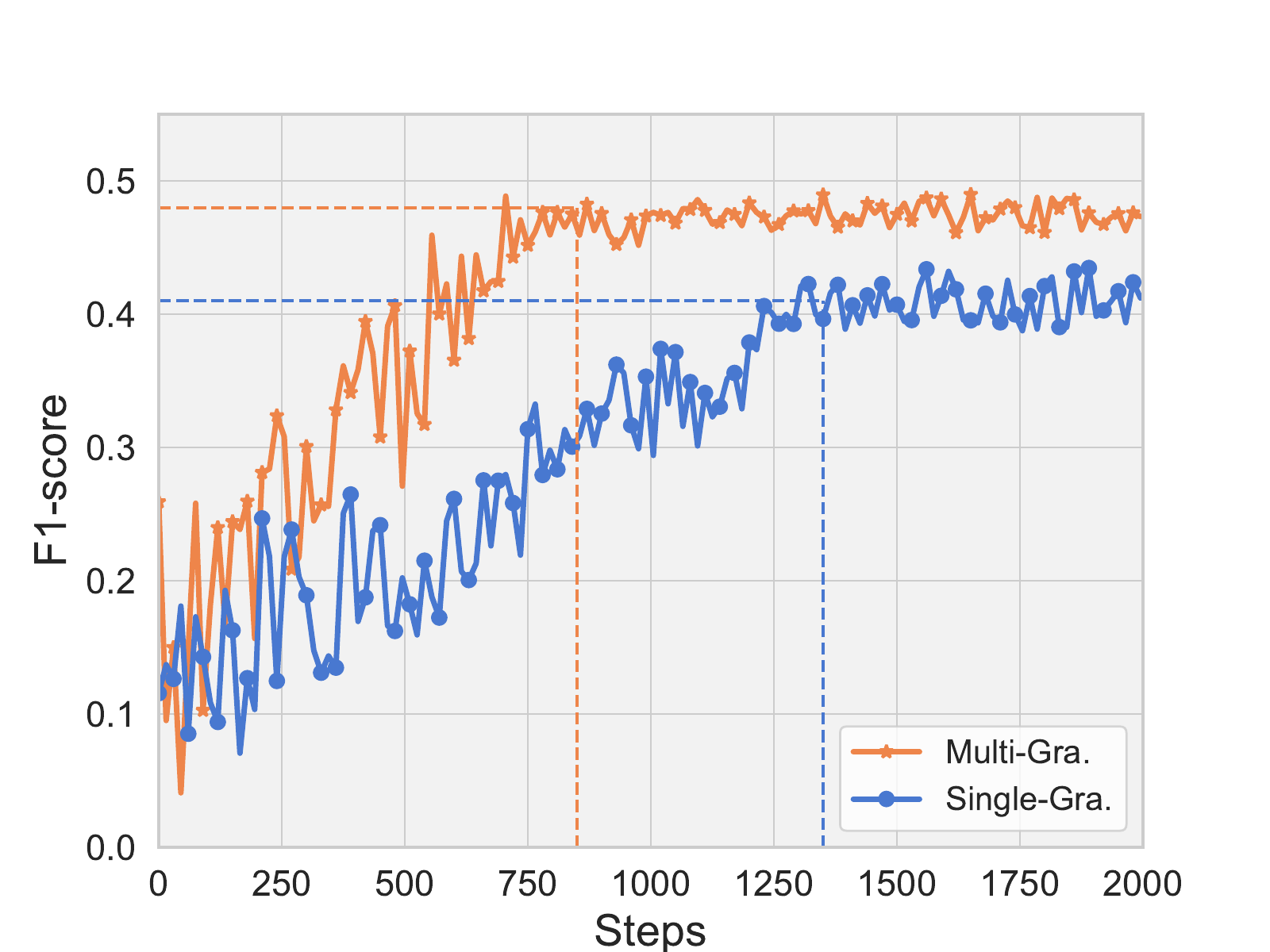} 
\caption{The comparison including convergence speed (i.e. steps) and performance (i.e. F1-score) of training process between our multi-granularity cross-level attention (Multi-Gra.) and single-granularity (Single-Gra.). (Best viewed in color)}
\label{multi_single}
\end{figure}

Figure \ref{multi_single} shows the comparison of final stable results and speed of convergence between the multi-granularity recontextualization and single-granularity\footnote{The single-granularity means just facility original representation of each level without combining the multi-headed attention mechanism.} on the NYT10 dataset.
we can observe that (1) multi-granularity recontextualization converges faster to not only stable but also better results.
(2) When the final performance of the model converge stably, our multi-granularity recontextualization have less jitter amplitude making our model more robust.

\begin{figure}[h]
\centering
\includegraphics[width=1.0\columnwidth]{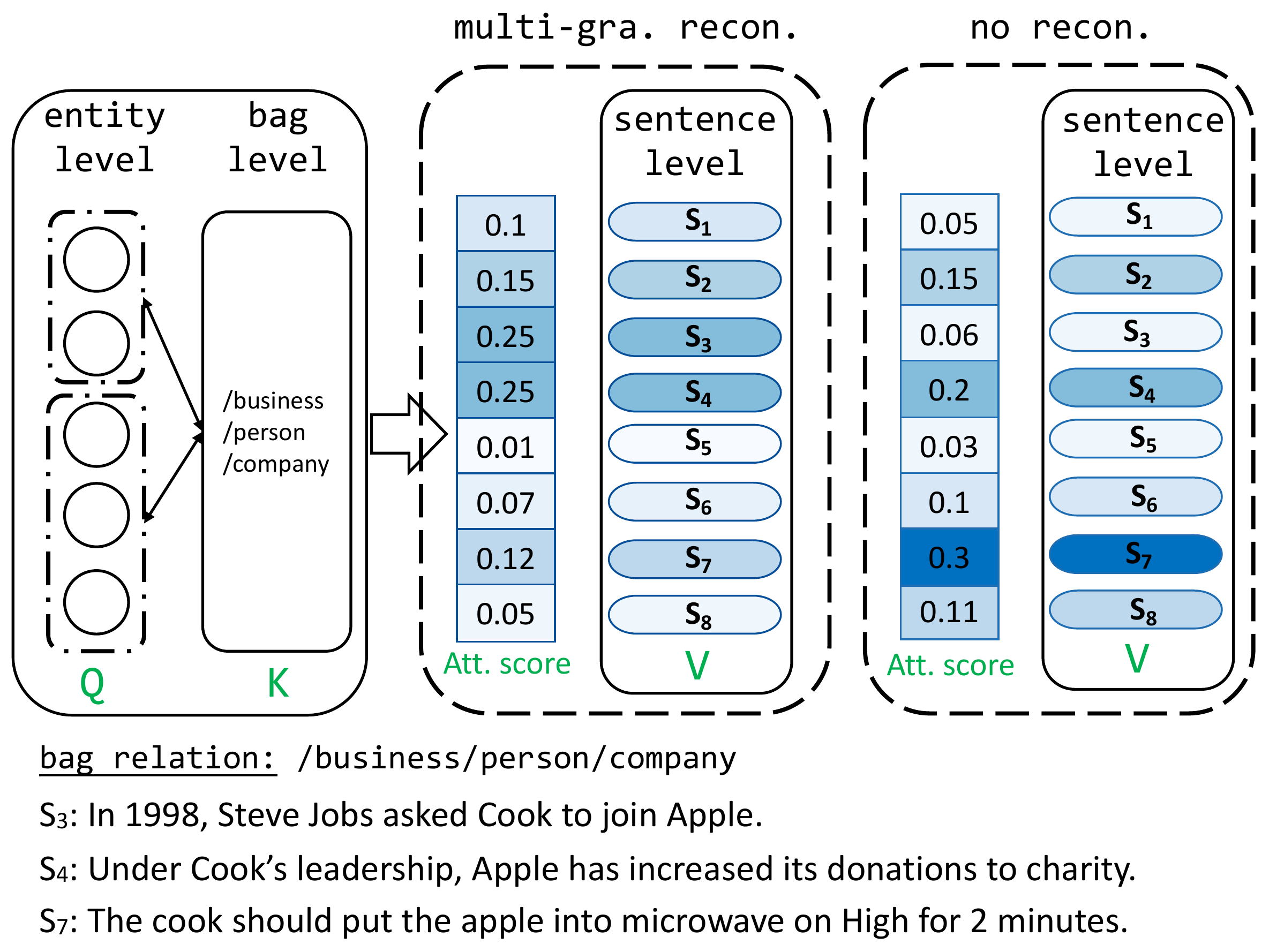} 
\caption{The heat maps of our multi-headed attention mechanism (ours) and no attention mechanism (right) among cross levels. (Best viewed in color)}
\label{heatmap}
\end{figure}

Figure \ref{heatmap} shows the two heat maps of the module with and without attention calculations, proving our multi-granularity recontextualization mechanism is effective for denoising the redundant sentences.
For example, we take the sentence-level representations act as ``V'' value.
Our multi-granularity recontextualization mechanism can achieve the higher attention scores (e.g. $S_3$ and $S_4$) for the important sentences in a bag, whereas the no recontextualization models incorrectly assign the highest attention score (e.g. $S_7$) to the noisy sentences.
This phenomenon indicates that this mechanism has a better ability to filter noisy sentences.


\subsection{The Influence of Gradient-based Data Augmentation}
To further prove our data augmentation skill of contrastive learning is effective, we choose the other three strategies (i.e. randomly deleting a word, twice dropout, and randomly noise) to perform the evolution process of representation learning space.

\begin{figure}[!ht]
\centering
\includegraphics[width=0.9\columnwidth]{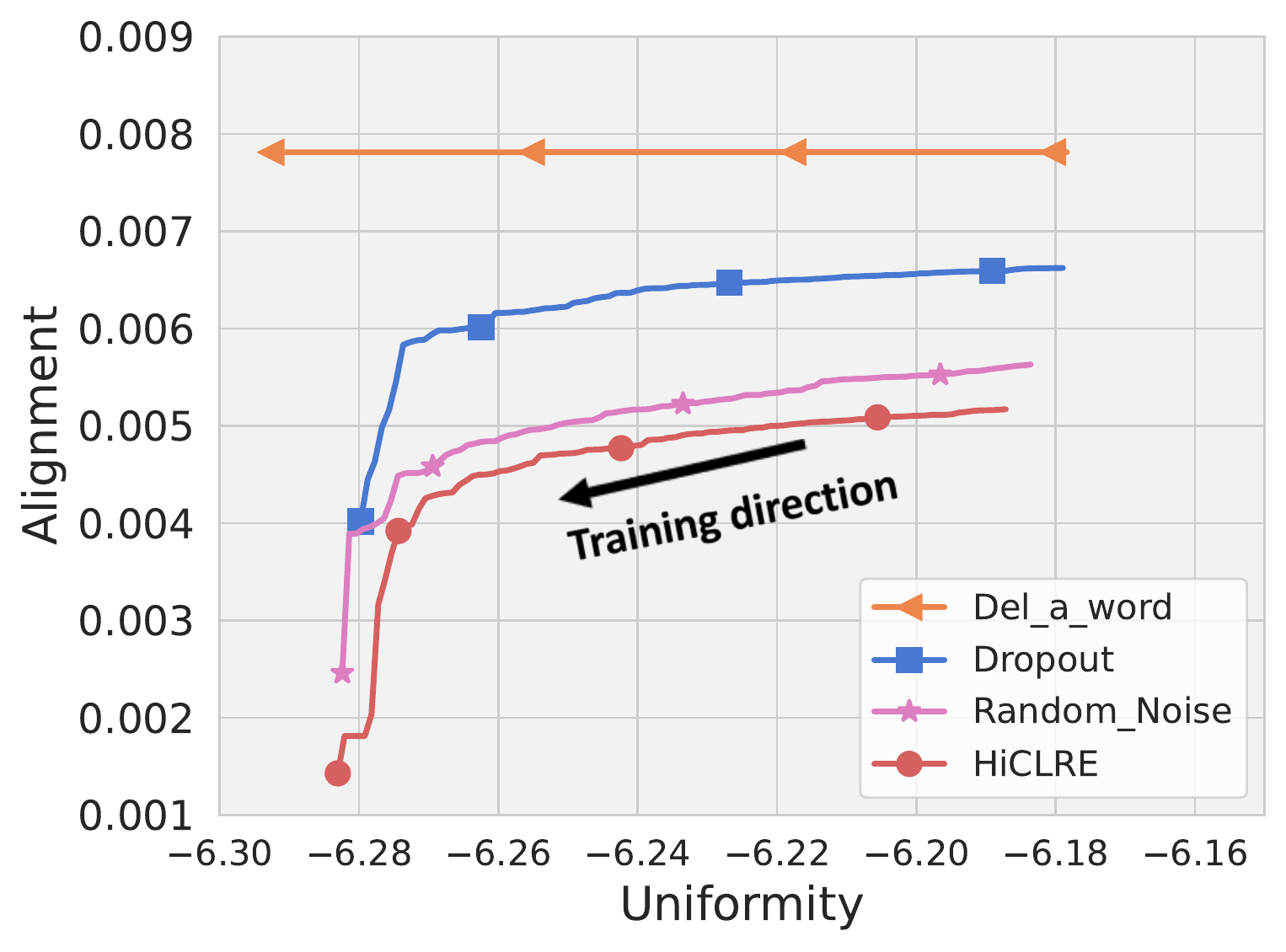} 
\caption{Results comparison of HiCLRE and other data augmentation skills in terms of alignment and uniformity. The arrows indicate the training direction. (Best viewed in color).}.
\vspace{-0.5cm}
\label{align_uniform}
\end{figure}

Following the previous works \citep{DBLP:conf/icml/0001I20}, we treat the pseudo sample as the positive instance and a randomly chosen instance from the batch as the negative instance to calculate alignment and uniformity. Then, we plot the transformation of the align-uniform points in Figure \ref{align_uniform}. 
The lower alignment and uniformity results indicate the better context-aware representations of the contrastive learning process.
Compared to the other positive sample generation skills, our dynamic gradient adversarial perturbation module reduces the alignment and uniformity metrics steadily to the lower value and faster speed.

\section{Conclusion}
In this paper, we propose HiCLRE, a hierarchical contrastive learning framework for distantly supervised relation extraction. 
Multi-Granularity Recontextualization module of HiCLRE utilizes a multi-head self-attention mechanism to transmit the information across three levels.
Dynamic Gradient Adversarial Perturbation module combines the gradient perturbation with inertia memory information to construct better pseudo positive samples for contrastive learning.
Experiments show the effectiveness of HiCLRE against the strong baseline models in various DSRE datasets.
\section*{Acknowledgements}
We would like to thank anonymous reviewers for their valuable comments.

\bibliography{anthology}

\begin{thebibliography}{49}
\expandafter\ifx\csname natexlab\endcsname\relax\def\natexlab#1{#1}\fi

\bibitem[{Alt et~al.(2020)Alt, Gabryszak, and Hennig}]{DBLP:conf/acl/AltGH20}
Christoph Alt, Aleksandra Gabryszak, and Leonhard Hennig. 2020.
\newblock \href {https://doi.org/10.18653/v1/2020.acl-main.140} {Probing
  linguistic features of sentence-level representations in relation
  extraction}.
\newblock In \emph{ACL}, pages 1534--1545.

\bibitem[{Alt et~al.(2019)Alt, H{\"{u}}bner, and
  Hennig}]{DBLP:conf/acl/AltHH19}
Christoph Alt, Marc H{\"{u}}bner, and Leonhard Hennig. 2019.
\newblock \href {https://doi.org/10.18653/v1/p19-1134} {Fine-tuning pre-trained
  transformer language models to distantly supervised relation extraction}.
\newblock In \emph{ACL}, pages 1388--1398.

\bibitem[{Beltagy et~al.(2019)Beltagy, Lo, and
  Ammar}]{DBLP:conf/naacl/BeltagyLA19}
Iz~Beltagy, Kyle Lo, and Waleed Ammar. 2019.
\newblock \href {https://doi.org/10.18653/v1/n19-1184} {Combining distant and
  direct supervision for neural relation extraction}.
\newblock In \emph{NAACL}, pages 1858--1867.

\bibitem[{Chen et~al.(2021{\natexlab{a}})Chen, Shi, Tang, Chen, Wu, and
  Zhuang}]{DBLP:conf/acl/ChenST00Z20}
Tao Chen, Haizhou Shi, Siliang Tang, Zhigang Chen, Fei Wu, and Yueting Zhuang.
  2021{\natexlab{a}}.
\newblock \href {https://doi.org/10.18653/v1/2021.acl-long.483} {{CIL:}
  contrastive instance learning framework for distantly supervised relation
  extraction}.
\newblock In \emph{ACL}, pages 6191--6200.

\bibitem[{Chen et~al.(2021{\natexlab{b}})Chen, Shi, Liu, Tang, Shao, Chen, and
  Zhuang}]{DBLP:conf/aaai/ChenSLTSCZ21}
Tao Chen, Haochen Shi, Liyuan Liu, Siliang Tang, Jian Shao, Zhigang Chen, and
  Yueting Zhuang. 2021{\natexlab{b}}.
\newblock \href {https://ojs.aaai.org/index.php/AAAI/article/view/17501}
  {Empower distantly supervised relation extraction with collaborative
  adversarial training}.
\newblock In \emph{AAAI}, pages 12675--12682.

\bibitem[{Christopoulou et~al.(2021)Christopoulou, Miwa, and
  Ananiadou}]{DBLP:conf/naacl/ChristopoulouMA21}
Fenia Christopoulou, Makoto Miwa, and Sophia Ananiadou. 2021.
\newblock \href {https://doi.org/10.18653/v1/2021.naacl-main.2} {Distantly
  supervised relation extraction with sentence reconstruction and knowledge
  base priors}.
\newblock In \emph{NAACL}, pages 11--26.

\bibitem[{Christou and Tsoumakas(2021)}]{DBLP:journals/access/ChristouT21}
Despina Christou and Grigorios Tsoumakas. 2021.
\newblock \href {https://doi.org/10.1109/ACCESS.2021.3073428} {Improving
  distantly-supervised relation extraction through bert-based label and
  instance embeddings}.
\newblock \emph{{IEEE} Access}, pages 62574--62582.

\bibitem[{Devlin et~al.(2019)Devlin, Chang, Lee, and
  Toutanova}]{DBLP:conf/naacl/DevlinCLT19}
Jacob Devlin, Ming{-}Wei Chang, Kenton Lee, and Kristina Toutanova. 2019.
\newblock \href {https://doi.org/10.18653/v1/n19-1423} {{BERT:} pre-training of
  deep bidirectional transformers for language understanding}.
\newblock In \emph{NAACL}, pages 4171--4186.

\bibitem[{Gao et~al.(2021)Gao, Yao, and Chen}]{DBLP:conf/emnlp/GaoYC21}
Tianyu Gao, Xingcheng Yao, and Danqi Chen. 2021.
\newblock \href {https://aclanthology.org/2021.emnlp-main.552} {Simcse: Simple
  contrastive learning of sentence embeddings}.
\newblock In \emph{EMNLP}, pages 6894--6910.

\bibitem[{Gutmann and Hyv{\"{a}}rinen(2010)}]{DBLP:journals/jmlr/GutmannH10}
Michael Gutmann and Aapo Hyv{\"{a}}rinen. 2010.
\newblock \href {http://proceedings.mlr.press/v9/gutmann10a.html}
  {Noise-contrastive estimation: {A} new estimation principle for unnormalized
  statistical models}.
\newblock In \emph{AISTATS}, pages 297--304.

\bibitem[{Hadsell et~al.(2006)Hadsell, Chopra, and
  LeCun}]{DBLP:conf/cvpr/HadsellCL06}
Raia Hadsell, Sumit Chopra, and Yann LeCun. 2006.
\newblock \href {https://doi.org/10.1109/CVPR.2006.100} {Dimensionality
  reduction by learning an invariant mapping}.
\newblock In \emph{CVPR}, pages 1735--1742.

\bibitem[{Hoffmann et~al.(2011)Hoffmann, Zhang, Ling, Zettlemoyer, and
  Weld}]{DBLP:conf/acl/HoffmannZLZW11}
Raphael Hoffmann, Congle Zhang, Xiao Ling, Luke~S. Zettlemoyer, and Daniel~S.
  Weld. 2011.
\newblock \href {https://aclanthology.org/P11-1055/} {Knowledge-based weak
  supervision for information extraction of overlapping relations}.
\newblock In \emph{ACL}, pages 541--550.

\bibitem[{Jaiswal et~al.(2020)Jaiswal, Babu, Zadeh, Banerjee, and
  Makedon}]{DBLP:journals/corr/abs-2011-00362}
Ashish Jaiswal, Ashwin~Ramesh Babu, Mohammad~Zaki Zadeh, Debapriya Banerjee,
  and Fillia Makedon. 2020.
\newblock \href {http://arxiv.org/abs/2011.00362} {A survey on contrastive
  self-supervised learning}.
\newblock \emph{CoRR}, abs/2011.00362.

\bibitem[{Jat et~al.(2017)Jat, Khandelwal, and
  Talukdar}]{DBLP:conf/akbc/JatKT17}
Sharmistha Jat, Siddhesh Khandelwal, and Partha~P. Talukdar. 2017.
\newblock \href {http://arxiv.org/abs/1804.06987} {Improving distantly
  supervised relation extraction using word and entity based attention}.
\newblock In \emph{NIPS}.

\bibitem[{Ji et~al.(2010)Ji, Grishman, Dang, Griffitt, and
  Ellis}]{ji2010overview}
Heng Ji, Ralph Grishman, Hoa~Trang Dang, Kira Griffitt, and Joe Ellis. 2010.
\newblock Overview of the tac 2010 knowledge base population track.
\newblock In \emph{TAC}, volume~3, pages 3--3.

\bibitem[{Khatib et~al.(2020)Khatib, Hou, Wachsmuth, Jochim, Bonin, and
  Stein}]{DBLP:conf/aaai/KhatibHWJB020}
Khalid~Al Khatib, Yufang Hou, Henning Wachsmuth, Charles Jochim, Francesca
  Bonin, and Benno Stein. 2020.
\newblock \href {https://aaai.org/ojs/index.php/AAAI/article/view/6231}
  {End-to-end argumentation knowledge graph construction}.
\newblock In \emph{AAAI}, pages 7367--7374.

\bibitem[{Li et~al.(2019)Li, Zhang, Jia, and Zhao}]{DBLP:conf/naacl/LiZJZ19}
Pengshuai Li, Xinsong Zhang, Weijia Jia, and Hai Zhao. 2019.
\newblock \href {https://doi.org/10.18653/v1/n19-1307} {{GAN} driven
  semi-distant supervision for relation extraction}.
\newblock In \emph{NAACL}, pages 3026--3035.

\bibitem[{Lin et~al.(2016)Lin, Shen, Liu, Luan, and
  Sun}]{DBLP:conf/acl/LinSLLS16}
Yankai Lin, Shiqi Shen, Zhiyuan Liu, Huanbo Luan, and Maosong Sun. 2016.
\newblock \href {https://doi.org/10.18653/v1/p16-1200} {Neural relation
  extraction with selective attention over instances}.
\newblock In \emph{ACL}.

\bibitem[{Ling and Weld(2012)}]{DBLP:conf/aaai/LingW12}
Xiao Ling and Daniel~S. Weld. 2012.
\newblock \href {http://www.aaai.org/ocs/index.php/AAAI/AAAI12/paper/view/5152}
  {Fine-grained entity recognition}.
\newblock In \emph{AAAI}.

\bibitem[{Liu et~al.(2020)Liu, Gong, Fu, Yan, Chen, Jiang, Lv, and
  Duan}]{DBLP:conf/acl/LiuGFYCJLD20}
Dayiheng Liu, Yeyun Gong, Jie Fu, Yu~Yan, Jiusheng Chen, Daxin Jiang, Jiancheng
  Lv, and Nan Duan. 2020.
\newblock \href {https://doi.org/10.18653/v1/2020.acl-main.604} {Rikinet:
  Reading wikipedia pages for natural question answering}.
\newblock In \emph{ACL}, pages 6762--6771.

\bibitem[{Loshchilov and Hutter(2017)}]{DBLP:journals/corr/abs-1711-05101}
Ilya Loshchilov and Frank Hutter. 2017.
\newblock \href {http://arxiv.org/abs/1711.05101} {Fixing weight decay
  regularization in adam}.
\newblock \emph{CoRR}.

\bibitem[{Ma et~al.(2021)Ma, Gui, Li, Zhang, Huang, and
  Zhou}]{DBLP:conf/acl/MaGLZHZ20}
Ruotian Ma, Tao Gui, Linyang Li, Qi~Zhang, Xuanjing Huang, and Yaqian Zhou.
  2021.
\newblock \href {https://doi.org/10.18653/v1/2021.acl-long.484} {{SENT:}
  sentence-level distant relation extraction via negative training}.
\newblock In \emph{ACL}, pages 6201--6213.

\bibitem[{Mintz et~al.(2009)Mintz, Bills, Snow, and
  Jurafsky}]{DBLP:conf/acl/MintzBSJ09}
Mike Mintz, Steven Bills, Rion Snow, and Daniel Jurafsky. 2009.
\newblock \href {https://aclanthology.org/P09-1113/} {Distant supervision for
  relation extraction without labeled data}.
\newblock In \emph{ACL}, pages 1003--1011.

\bibitem[{Qin et~al.(2018)Qin, Xu, and Wang}]{DBLP:conf/acl/WangXQ18}
Pengda Qin, Weiran Xu, and William~Yang Wang. 2018.
\newblock \href {https://doi.org/10.18653/v1/P18-1046} {{DSGAN:} generative
  adversarial training for distant supervision relation extraction}.
\newblock In \emph{ACL}, pages 496--505.

\bibitem[{Qin et~al.(2021)Qin, Lin, Takanobu, Liu, Li, Ji, Huang, Sun, and
  Zhou}]{DBLP:conf/acl/QinLT00JHS020}
Yujia Qin, Yankai Lin, Ryuichi Takanobu, Zhiyuan Liu, Peng Li, Heng Ji, Minlie
  Huang, Maosong Sun, and Jie Zhou. 2021.
\newblock \href {https://doi.org/10.18653/v1/2021.acl-long.260} {{ERICA:}
  improving entity and relation understanding for pre-trained language models
  via contrastive learning}.
\newblock In \emph{ACL}, pages 3350--3363.

\bibitem[{Riedel et~al.(2010)Riedel, Yao, and
  McCallum}]{DBLP:conf/pkdd/RiedelYM10}
Sebastian Riedel, Limin Yao, and Andrew McCallum. 2010.
\newblock \href {https://doi.org/10.1007/978-3-642-15939-8\_10} {Modeling
  relations and their mentions without labeled text}.
\newblock In \emph{ECML}, pages 148--163.

\bibitem[{Saxena et~al.(2020)Saxena, Tripathi, and
  Talukdar}]{DBLP:conf/acl/SaxenaTT20}
Apoorv Saxena, Aditay Tripathi, and Partha~P. Talukdar. 2020.
\newblock \href {https://doi.org/10.18653/v1/2020.acl-main.412} {Improving
  multi-hop question answering over knowledge graphs using knowledge base
  embeddings}.
\newblock In \emph{ACL}, pages 4498--4507.

\bibitem[{Shi and Eberhart(1998)}]{DBLP:conf/eps/ShiE98}
Yuhui Shi and Russell~C. Eberhart. 1998.
\newblock \href {https://doi.org/10.1007/BFb0040810} {Parameter selection in
  particle swarm optimization}.
\newblock In \emph{Evolutionary Programming}, pages 591--600.

\bibitem[{Soares et~al.(2019)Soares, FitzGerald, Ling, and
  Kwiatkowski}]{DBLP:conf/acl/SoaresFLK19}
Livio~Baldini Soares, Nicholas FitzGerald, Jeffrey Ling, and Tom Kwiatkowski.
  2019.
\newblock \href {https://doi.org/10.18653/v1/p19-1279} {Matching the blanks:
  Distributional similarity for relation learning}.
\newblock In \emph{ACL}, pages 2895--2905.

\bibitem[{Su et~al.(2018)Su, Jia, Cheng, Zhu, and
  Li}]{DBLP:conf/ijcai/SuJ0ZL18}
Sen Su, Ningning Jia, Xiang Cheng, Shuguang Zhu, and Ruiping Li. 2018.
\newblock \href {https://doi.org/10.24963/ijcai.2018/610} {Exploring
  encoder-decoder model for distant supervised relation extraction}.
\newblock In \emph{IJCAI}, pages 4389--4395.

\bibitem[{Surdeanu et~al.(2012)Surdeanu, Tibshirani, Nallapati, and
  Manning}]{DBLP:conf/emnlp/SurdeanuTNM12}
Mihai Surdeanu, Julie Tibshirani, Ramesh Nallapati, and Christopher~D. Manning.
  2012.
\newblock \href {https://aclanthology.org/D12-1042/} {Multi-instance
  multi-label learning for relation extraction}.
\newblock In \emph{EMNLP}, pages 455--465.

\bibitem[{Tang et~al.(2020)Tang, Huang, Wang, He, and
  Zhou}]{DBLP:conf/acl/TangHWHZ20}
Yun Tang, Jing Huang, Guangtao Wang, Xiaodong He, and Bowen Zhou. 2020.
\newblock \href {https://doi.org/10.18653/v1/2020.acl-main.241} {Orthogonal
  relation transforms with graph context modeling for knowledge graph
  embedding}.
\newblock In \emph{ACL}, pages 2713--2722.

\bibitem[{van~den Oord et~al.(2018)van~den Oord, Li, and
  Vinyals}]{DBLP:journals/corr/abs-1807-03748}
A{\"{a}}ron van~den Oord, Yazhe Li, and Oriol Vinyals. 2018.
\newblock \href {http://arxiv.org/abs/1807.03748} {Representation learning with
  contrastive predictive coding}.
\newblock \emph{CoRR}.

\bibitem[{van~der Maaten and Hinton(2008)}]{van2008visualizing}
Laurens van~der Maaten and Geoffrey Hinton. 2008.
\newblock Visualizing data using t-sne.
\newblock pages 2579--2605.

\bibitem[{Vashishth et~al.(2018)Vashishth, Joshi, Prayaga, Bhattacharyya, and
  Talukdar}]{DBLP:conf/emnlp/VashishthJPBT18}
Shikhar Vashishth, Rishabh Joshi, Sai~Suman Prayaga, Chiranjib Bhattacharyya,
  and Partha~P. Talukdar. 2018.
\newblock \href {https://doi.org/10.18653/v1/d18-1157} {{RESIDE:} improving
  distantly-supervised neural relation extraction using side information}.
\newblock In \emph{EMNLP}, pages 1257--1266.

\bibitem[{Vaswani et~al.(2017)Vaswani, Shazeer, Parmar, Uszkoreit, Jones,
  Gomez, Kaiser, and Polosukhin}]{DBLP:conf/nips/VaswaniSPUJGKP17}
Ashish Vaswani, Noam Shazeer, Niki Parmar, Jakob Uszkoreit, Llion Jones,
  Aidan~N. Gomez, Lukasz Kaiser, and Illia Polosukhin. 2017.
\newblock \href
  {https://proceedings.neurips.cc/paper/2017/hash/3f5ee243547dee91fbd053c1c4a845aa-Abstract.html}
  {Attention is all you need}.
\newblock In \emph{NIPS}, pages 5998--6008.

\bibitem[{Veyseh et~al.(2020)Veyseh, Dernoncourt, Dou, and
  Nguyen}]{DBLP:conf/acl/VeysehDDN20}
Amir Pouran~Ben Veyseh, Franck Dernoncourt, Dejing Dou, and Thien~Huu Nguyen.
  2020.
\newblock \href {https://doi.org/10.18653/v1/2020.acl-main.715} {Exploiting the
  syntax-model consistency for neural relation extraction}.
\newblock In \emph{ACL}, pages 8021--8032.

\bibitem[{Wang and Jiang(2019)}]{DBLP:conf/acl/WangJ19}
Chao Wang and Hui Jiang. 2019.
\newblock \href {https://doi.org/10.18653/v1/p19-1219} {Explicit utilization of
  general knowledge in machine reading comprehension}.
\newblock In \emph{ACL}, pages 2263--2272.

\bibitem[{Wang and Isola(2020)}]{DBLP:conf/icml/0001I20}
Tongzhou Wang and Phillip Isola. 2020.
\newblock \href {http://proceedings.mlr.press/v119/wang20k.html} {Understanding
  contrastive representation learning through alignment and uniformity on the
  hypersphere}.
\newblock In \emph{ICML}, pages 9929--9939.

\bibitem[{Wei and Zou(2019)}]{DBLP:conf/emnlp/WeiZ19}
Jason~W. Wei and Kai Zou. 2019.
\newblock \href {https://doi.org/10.18653/v1/D19-1670} {{EDA:} easy data
  augmentation techniques for boosting performance on text classification
  tasks}.
\newblock In \emph{EMNLP}, pages 6381--6387.

\bibitem[{Wei et~al.(2020)Wei, Su, Wang, Tian, and
  Chang}]{DBLP:conf/acl/WeiSWTC20}
Zhepei Wei, Jianlin Su, Yue Wang, Yuan Tian, and Yi~Chang. 2020.
\newblock \href {https://doi.org/10.18653/v1/2020.acl-main.136} {A novel
  cascade binary tagging framework for relational triple extraction}.
\newblock In \emph{ACL}, pages 1476--1488.

\bibitem[{Wu et~al.(2019)Wu, Fan, and Zhang}]{DBLP:conf/aaai/WuFZ19}
Shanchan Wu, Kai Fan, and Qiong Zhang. 2019.
\newblock \href {https://doi.org/10.1609/aaai.v33i01.33017273} {Improving
  distantly supervised relation extraction with neural noise converter and
  conditional optimal selector}.
\newblock In \emph{AAAI}, pages 7273--7280.

\bibitem[{Yan et~al.(2021)Yan, Li, Wang, Zhang, Wu, and
  Xu}]{DBLP:conf/acl/YanLWZWX20}
Yuanmeng Yan, Rumei Li, Sirui Wang, Fuzheng Zhang, Wei Wu, and Weiran Xu. 2021.
\newblock \href {https://doi.org/10.18653/v1/2021.acl-long.393} {Consert: {A}
  contrastive framework for self-supervised sentence representation transfer}.
\newblock In \emph{ACL}, pages 5065--5075.

\bibitem[{Yao et~al.(2011)Yao, Haghighi, Riedel, and
  McCallum}]{DBLP:conf/emnlp/YaoHRM11}
Limin Yao, Aria Haghighi, Sebastian Riedel, and Andrew McCallum. 2011.
\newblock \href {https://aclanthology.org/D11-1135/} {Structured relation
  discovery using generative models}.
\newblock In \emph{EMNLP}, pages 1456--1466.

\bibitem[{Ye and Ling(2019)}]{DBLP:conf/naacl/YeL19}
Zhi{-}Xiu Ye and Zhen{-}Hua Ling. 2019.
\newblock \href {https://doi.org/10.18653/v1/n19-1288} {Distant supervision
  relation extraction with intra-bag and inter-bag attentions}.
\newblock In \emph{NAACL}, pages 2810--2819.

\bibitem[{Zang et~al.(2020)Zang, Qi, Yang, Liu, Zhang, Liu, and
  Sun}]{DBLP:conf/acl/ZangQYLZLS20}
Yuan Zang, Fanchao Qi, Chenghao Yang, Zhiyuan Liu, Meng Zhang, Qun Liu, and
  Maosong Sun. 2020.
\newblock \href {https://doi.org/10.18653/v1/2020.acl-main.540} {Word-level
  textual adversarial attacking as combinatorial optimization}.
\newblock In \emph{ACL}, pages 6066--6080.

\bibitem[{Zeng et~al.(2015)Zeng, Liu, Chen, and
  Zhao}]{DBLP:conf/emnlp/ZengLC015}
Daojian Zeng, Kang Liu, Yubo Chen, and Jun Zhao. 2015.
\newblock \href {https://doi.org/10.18653/v1/d15-1203} {Distant supervision for
  relation extraction via piecewise convolutional neural networks}.
\newblock In \emph{EMNLP}, pages 1753--1762.

\bibitem[{Zhang et~al.(2020)Zhang, Sheng, Alhazmi, and
  Li}]{DBLP:journals/tist/ZhangSAL20}
Wei~Emma Zhang, Quan~Z. Sheng, Ahoud Abdulrahmn~F. Alhazmi, and Chenliang Li.
  2020.
\newblock \href {https://doi.org/10.1145/3374217} {Adversarial attacks on
  deep-learning models in natural language processing: {A} survey}.
\newblock \emph{ACM}, pages 24:1--24:41.

\bibitem[{Zou et~al.(2020)Zou, Huang, Xie, Dai, and
  Chen}]{DBLP:conf/acl/ZouHXDC20}
Wei Zou, Shujian Huang, Jun Xie, Xinyu Dai, and Jiajun Chen. 2020.
\newblock \href {https://doi.org/10.18653/v1/2020.acl-main.319} {A reinforced
  generation of adversarial examples for neural machine translation}.
\newblock In \emph{ACL}, pages 3486--3497.

\end{thebibliography}
\bibliographystyle{acl_natbib}

\newpage
\appendix

\section{Datasets Statistics}
\label{appendix_Datasets_Statistics}

\begin{table}[!ht]
\footnotesize
\centering
\begin{tabular}{c|r|r|r|c}
\toprule
Dataset & \# Rel. & \# Train & \# Test & Test Type \\ \midrule
NYT10   & 58     & 522,611   & 172,448  & DS        \\
GDS     & 5      & 18,328    & 5,663    & Partly MA \\
NYT10-M & 25     & 417,893   & 11,085   & MA        \\
KBP     & 12     & 87,940    & 288     & MA        \\ \bottomrule
\end{tabular}
\caption{Statistics of four datasets. Rel.: relation, DS: distantly supervised and MA: manually annotated.}
\label{dataset_details}
\end{table}

\begin{figure*}[ht]
\centering
\includegraphics[width=2.0\columnwidth]{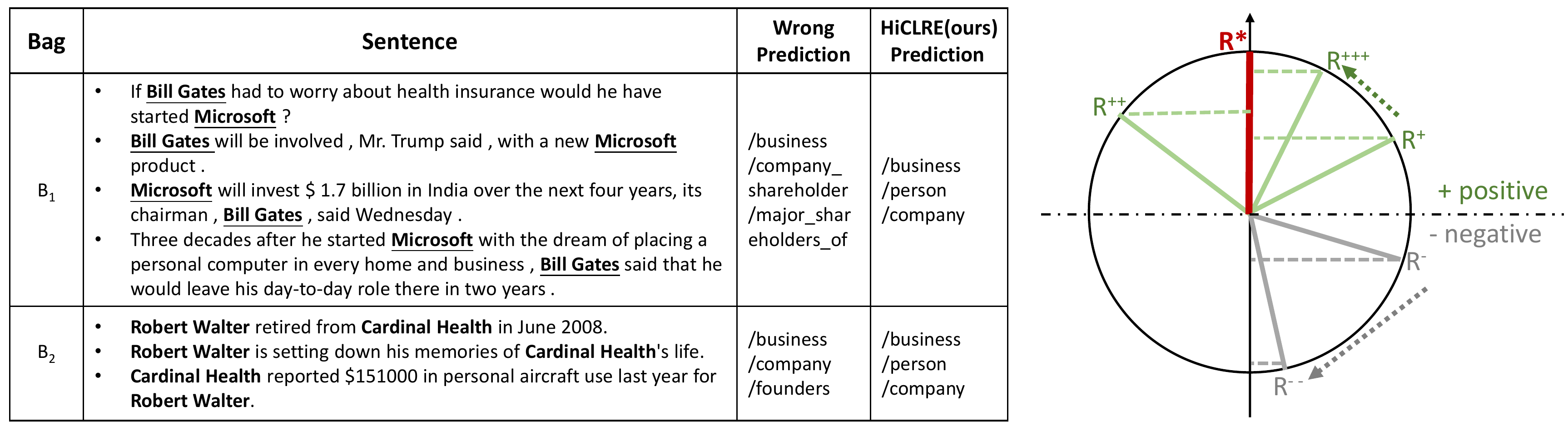} 
\caption{Examples of cases in our experiments. The left table means the comparison of predicted labels. The right figure shows the Dynamic Gradient Adversarial Perturbation module's working process.(Best viewed in color).}.
\vspace{-1em}
\label{case_study}
\end{figure*}
\section{Case Study}
\label{appendix_Case_Study}
We enumerate several representative examples in Figure \ref{case_study} to further explore why our model can work in the distantly supervised scenario.
In the left part of the figure, there are two bags containing the different entity pairs (i.e. $\langle$``Bill Gates'', ``Microsoft'' $\rangle$ and $\langle$``Robert Walter'', ``Cardinal Health'' $\rangle$).
Previous works ignore the consideration of the representation interaction in a specific levels and cross levels, which may be hard to predict similar or difficult instances. For example, sentences of bag $B_{1}$ are always classified into the relation ``\textit{major\_shareholders\_of}''. 
Although these two relations (i.e. ``\textit{major\_shareholders\_of}'' and  ``\textit{/person/company}'') are pretty similar to each other, none of the four sentences' semantics in $B_{1}$ represent the meaning of ``\textit{major\_shareholders\_of}''.
In particular, after the context-aware representations interaction via cross levels and specific levels, HiCLRE can correctly predict the bag to the ground-truth label; likewise, the bag $B_{2}$ is in the same situation.

In the right part of the figure, we demonstrate an example of that HiCLRE can pull the correlated instance closely and push the uncorrelated instance away. We reduce the dimension of bag examples' representations by t-SNE \cite{van2008visualizing} and show the example results in the coordinate system. $R^{*}$ is the target instance to be classified, the symbol ``$+$'' represents the degree of relevance and ``$-$'' represents the degree of irrelevance. HiCLRE can pull the related sample $R^{+}$ to $R^{+++}$ (i.e. closer to $R^{*}$), while pushing the uncorrelated sample $R^{-}$ to $R^{--}$ (farther from $R^{*}$). This phenomenon is own to the design of gradient-based perturbation, which gives significant enhancement to interactions in a specific level.

\section{The influence of important hyper parameters}
\label{appendix_parameters_commissioning_process}
We experiment with our model on the NYT10 dataset with four important hyper-parameters, and discover a suitable parameters' combination to reach better performance.
\begin{figure*}[ht]
\centering
\includegraphics[width=2.0\columnwidth]{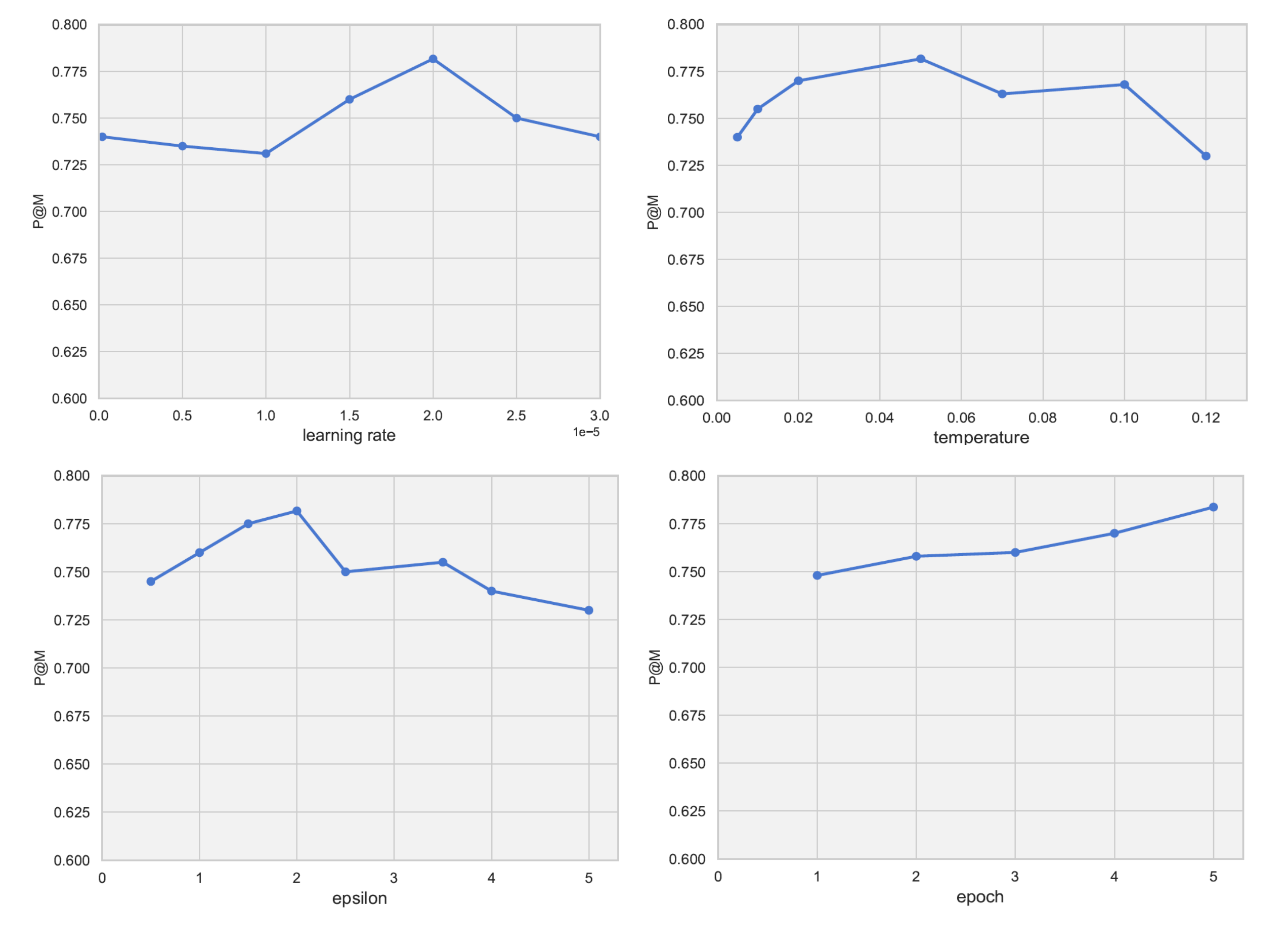} 
\caption{The influence of four important hyper-parameters on the NYT10 dataset. (Best viewed in color).}.
\vspace{-1em}
\label{HyperParameter}
\end{figure*}

\end{document}